\documentclass[runningheads]{llncs}
\usepackage[T1]{fontenc}
\usepackage{amsmath}
\usepackage{amssymb}
\usepackage{tabularx}

\usepackage{graphicx}
\begin{document}
\title{Enhancing Multimodal Emotion Recognition via Multi-Feature Encoding and Attention-Based Fusion}
\titlerunning{Multimodal Emotion Recognition with Multi-Feature Fusion}
%
%
\author{Xu Lin\inst{1}\thanks{Corresponding authors: Hui Kang and Xinying Wang} \and
Ke Wang\inst{1} \and
Hui Kang\inst{1,2} \and
Xinying Wang\inst{1,2}}

\authorrunning{X. Lin et al.}

\institute{College of Computer Science and Technology, Jilin University, Changchun 130012, China\\
\email{jlu\_linxu@163.com, iop6632@163.com, kanghui@jlu.edu.cn, xinying@jlu.edu.cn}
\and
Key Laboratory of Symbolic Computation and Knowledge Engineering of Ministry of Education, Jilin University, Changchun 130012, China}

\maketitle              
\begin{abstract}
    Multimodal emotion recognition has attracted growing interest 
    due to its importance in human-computer interaction, remote 
    education, and healthcare. This paper proposes a novel multimodal 
    emotion recognition framework that integrates rich audio and visual 
    feature extraction with an attention-based fusion strategy. For 
    audio, we extract three complementary feature types: semantic 
    embeddings from Wav2Vec2, MFCC features, and statistical acoustic 
    descriptors such as pitch, energy, and rhythm. These are aligned 
    and fused via a BiLSTM to capture temporal dependencies. For video, 
    we propose a ResNet50-BiLSTM architecture that combines deep residual 
    learning and sequential modeling to extract expressive spatiotemporal 
    features from facial sequences. To enhance multimodal synergy, we 
    introduce a feature-level fusion mechanism based on multi-head 
    attention, allowing the model to adaptively weigh contributions across 
    modalities. Experiments conducted on the MELD and IEMOCAP datasets 
    demonstrate that our model significantly outperforms baselines in both 
    accuracy and robustness.Furthermore, ablation studies show 
    that the attention-based fusion strategy significantly improves 
    performance in unbalanced data settings. Our findings suggest that 
    the proposed framework effectively captures diverse emotional cues 
    from speech and visual expressions, and offers a practical and 
    generalizable approach for real-world mutimodal emotion 
    recognition tasks.

\keywords{Emotion Recognition  \and Feature Extraction \and Multimodal Fusion \and Attention Mechanism.}
\end{abstract}
\section{Introduction}

\sloppy
Emotion plays a vital role in human communication, influencing decision-making, 
social interaction, and cognition~\cite{ref_article1}. As human-computer interaction 
systems become increasingly prevalent, enabling machines to accurately recognize 
and respond to human emotions is a key goal in affective computing~\cite{ref_article2}.

Conventional unimodal approaches often relied on single-modal data such as speech, 
facial expressions, or physiological signals. However, these approaches are susceptible to noise 
contamination and information scarcity, along with lacking contextual relevance, 
resulting in reduced robustness and biased emotional interpretation.

In fact, human emotion perception depends on multiple modalities~\cite{ref_lncs1}, 
integrating auditory, visual, and contextual cues. For instance, if a stressed friend replies 
``I'm doing great'' while looking frustrated, a unimodal speech-based system may incorrectly 
predict a positive emotion. Only by integrating visual and contextual signals can the true 
emotional state be recognized~\cite{ref_lncs2}.

Multimodal emotion recognition has advanced significantly. For example, Zhao et al. 
proposed VAANet~\cite{ref_lncs3}, integrating spatial, channel, and temporal attention 
mechanisms in a CNN. Avro et al. introduced EmoTech~\cite{ref_lncs4}, combining low-level 
audio and text features with a hybrid BiLSTM-CNN. However, challenges such as feature 
redundancy and insufficient cross-modal fusion remain~\cite{ref_lncs5}.

To address these, we propose a multimodal framework inspired by human perception. 
We design unimodal encoders: an audio module combining three types of acoustic features 
processed via BiLSTM, and a video module using ResNet-BiLSTM to capture spatiotemporal 
facial dynamics. A cross-modal attention mechanism then fuses the extracted features.

Experiments on MELD and IEMOCAP validate the effectiveness of our framework, 
which outperforms state-of-the-art methods in overall accuracy and minority-class performance.

\section{Methods}

Our multimodal emotion recognition framework comprises three key components: 
(1) a BiLSTM-based audio module that integrates and models temporal dependencies 
across three complementary audio features; 
(2) a ResNet50-BiLSTM video module for extracting deep spatiotemporal 
features from facial sequences; and 
(3) a multi-head attention-based fusion module that adaptively combines 
modalities at the feature level.

\subsection{BiLSTM-based Multi-Feature Audio Emotion Recognition}
\label{sec:audio}

\begin{figure}[h]
\includegraphics[width=0.98\textwidth]{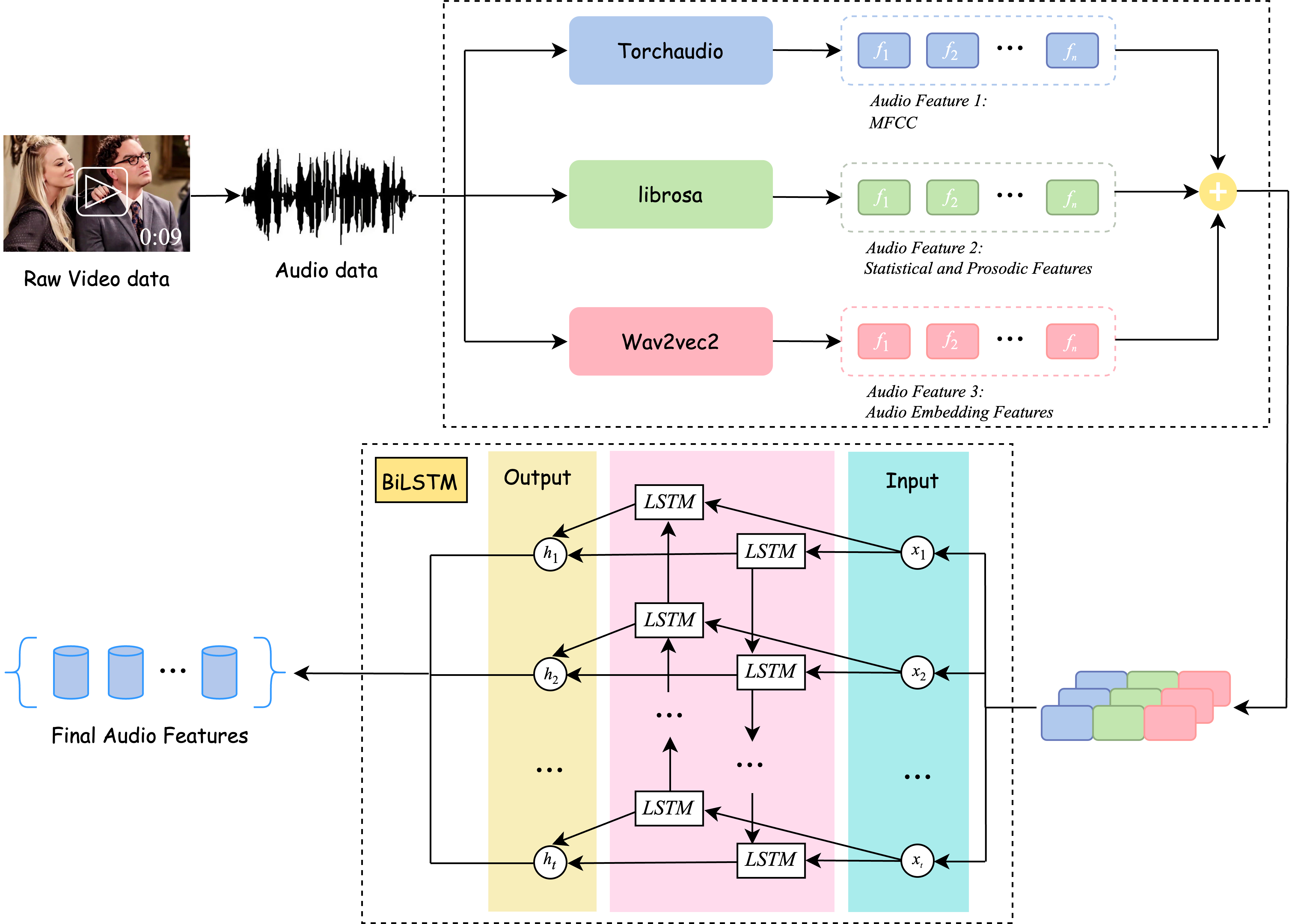}
\caption{Architecture of the audio feature extraction model } \label{fig1}
\end{figure}

The structure of the proposed BiLSTM-based audio emotion recognition module is 
illustrated in Fig.~\ref{fig1}. We first extract { \bfseries Mel-frequency cepstral 
coefficients (MFCCs)} using the Torchaudio library to capture low-level spectral 
features. Next, we employ the pre-trained Wav2Vec2 model to obtain 
{\bfseries semantic-level audio embeddings}. In parallel, we use the Librosa 
library to compute various prosodic and statistical descriptors, including 
{ \bfseries spectral centroid}, {\bfseries zero-crossing rate}, {\bfseries pitch}, 
{\bfseries root mean square (RMS) energy}, and {\bfseries rhythm}. Finally, these 
features are fed into a BiLSTM network for temporal sequence modeling and 
contextual information extraction across feature dimensions.
We note that the choice of libraries is intentional: Torchaudio is used for its 
seamless integration with the PyTorch ecosystem and GPU acceleration during MFCC 
extraction, which significantly improves preprocessing efficiency. Librosa, on 
the other hand, is employed for its comprehensive suite of hand-crafted prosodic 
and statistical descriptors beyond MFCCs. This combination leverages the respective 
strengths of each library.

\subsubsection{Multi-Feature Audio Feature Extraction Model}
\sloppy
Wav2Vec2 is a self-supervised model based on the Transformer architecture, consisting 
of a Feature Encoder and a Transformer Context module. The former maps each 20ms 
audio segment into a low-level feature vector, while the latter captures long-term 
dependencies and outputs semantic audio embeddings. The MFCC features approximate the human 
cochlea's nonlinear frequency response and describe the spectral envelope of speech signals. 
The statistical audio features extracted via Librosa provide explicit acoustic cues 
related to emotional expression.

These three types of features are complementary to each other, enabling 
deep exploration of the intrinsic structure of audio signals, and improving 
both the robustness and expressive power of the system.

\subsubsection{BiLSTM-Based Audio Feature Fusion Model}
Let the three types of extracted features be:
\begin{itemize}
    \item $\mathbf{V}_1$: Wav2Vec2-based semantic embedding (dimension $1 \times 512$)
    \item $\mathbf{V}_3$: MFCC features (dimension $1 \times 25$)
    \item $\mathbf{V}_4$: Statistical audio features (dimension $1 \times 25$)
\end{itemize}

We first project $\mathbf{V}_3$ and $\mathbf{V}_4$ to the same dimension as $\mathbf{V}_1$ via fully connected layers:
\begin{equation}
    \mathbf{V}_3' = \text{FC}_1(\mathbf{V}_3) , \quad
    \mathbf{V}_4' = \text{FC}_2(\mathbf{V}_4) 
\end{equation}

where $\mathbf{V}_3'\in \mathbb{R}^{1 \times 512}$,$\mathbf{V}_3'\in \mathbb{R}^{1 \times 512}$.
The three features are fused by weighted summation:
\begin{equation}
    \mathbf{F}_{\text{audio}} = \alpha \mathbf{V}_1 + \beta \mathbf{V}_3' + \gamma \mathbf{V}_4'
\end{equation}

where $\alpha+\beta+\gamma=1$.Then, the vector is reshaped to match the expected BiLSTM input format:
\begin{equation}
    \mathbf{F}_{\text{audio}}^{(t)} = \text{Unsqueeze}(\mathbf{F}_{\text{audio}}) 
\end{equation}

where $\mathbf{F}_{\text{audio}}^{(t)} \in \mathbb{R}^{1 \times 1 \times 512}$ .It is then passed through a bidirectional LSTM layer:
\begin{equation}
    \overrightarrow{\mathbf{h}} = \text{LSTM}_{\text{fwd}}(\mathbf{F}_{\text{audio}}^{(t)}), \quad
    \overleftarrow{\mathbf{h}} = \text{LSTM}_{\text{bwd}}(\mathbf{F}_{\text{audio}}^{(t)})
\end{equation}

The final audio representation is obtained by concatenating forward and backward outputs:
\begin{equation}
    \mathbf{z}_{\text{audio}} = [\overrightarrow{\mathbf{h}};\, \overleftarrow{\mathbf{h}}] 
\end{equation}

In this equation, $\mathbf{z}_{\text{audio}} \in \mathbb{R}^{1 \times 512}$

\subsection{BiLSTM-based Video Emotion Recognition Module}
\label{sec:video}

\begin{figure}[h]
\includegraphics[width=0.98\textwidth]{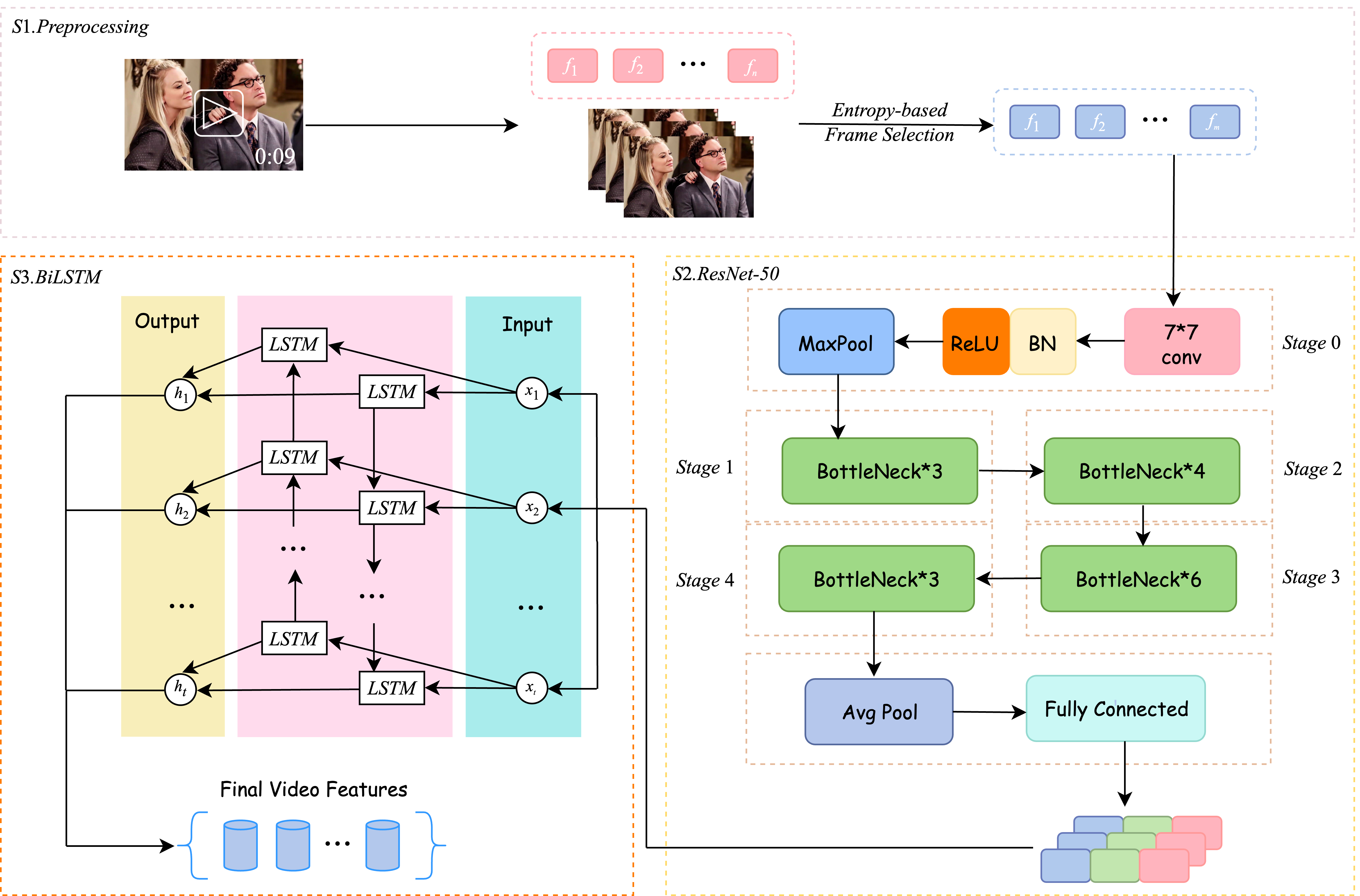}
\caption{Architecture of the video feature extraction model} \label{fig2}
\end{figure}

The architecture of the BiLSTM-based video emotion recognition module is illustrated in 
Fig.~\ref{fig2}. 

It consists primarily of a ResNet-50 backbone and a BiLSTM layer for temporal 
modeling.We first apply an entropy-based frame selection algorithm to extract a 
fixed number of representative frames from raw video data. These frames are 
selected either at uniform time intervals or based on content-aware criteria, 
ensuring that frames with the highest information content are retained.
The selected frames are then preprocessed and passed through a pre-trained 
ResNet-50 network, which extracts deep spatial features from each frame and 
outputs corresponding frame-level feature vectors.
These vectors are fed sequentially into a BiLSTM network to model temporal 
dependencies and refine the emotional context. By processing the sequence 
in both forward and backward directions, the BiLSTM captures bidirectional 
relationships and generates the final video-level feature representation.

\subsubsection{Entropy-Based Frame Selection} To reduce redundancy and focus on 
highly informative content, we applied an entropy-based frame selection algorithm 
to each video segment. Specifically, a fixed number of representative frames 
were extracted either at fixed time intervals or based on content-aware 
entropy scoring. This strategy ensures that the most informative frames 
are retained, thereby improving model efficiency and reducing noise 
interference.

\subsubsection{ResNet-50-Based Feature Extraction Model} 
The selected high-information-content frames are preprocessed and fed into a 
ResNet-50 network for high-level feature extraction. ResNet-50 employs residual 
units with skip connections, where the input is added directly to the output. 
This residual learning pathway facilitates gradient flow during backpropagation, 
mitigates vanishing gradients, and enables the training of deeper networks without
accuracy degradation. The formula for the residual module is:

\begin{equation}
    \boldsymbol{y}=\mathrm{H}(\boldsymbol{x},\mathbf{W_H})+\boldsymbol{x}
\end{equation}

Specifically, we first apply an initial convolutional layer 
to extract low-level features from the input image, expressed as:

\begin{equation}
    \mathbf{Y}=\mathrm{Conv2D}(\mathbf{X},\mathbf{W_1})+\boldsymbol{b_1}
\end{equation}

where $\mathbf{Y}$ is the output feature map, $\mathbf{X}$ is the input image,
$\mathbf{W_1}$ represents the $7 \times 7$ convolutional kernel weights with
stride 2 and 64 output channels, and $\boldsymbol{b_1}$ denotes the bias term.

Next, the output feature map is passed through a max pooling layer 
to reduce its spatial dimensions by selecting the maximum value in 
each local region:

\begin{equation}
    \mathbf{Z}=\mathrm{MaxPool2D}(\mathbf{Y})
\end{equation}

where $\mathbf{Z}$ is the output feature map and $\mathbf{Y}$ is the input feature map.

ResNet-50 is composed of four stages, containing 3, 4, 6, and 3 
residual blocks respectively. Each block employs skip connections 
that span two to three convolutional layers, establishing direct 
gradient flow pathways and improving the efficiency of parameter 
updates. This design helps mitigate the vanishing gradient problem 
in deep networks. The residual block is computed as:

\begin{equation}
    F(\mathbf{X})=\mathrm{Conv2D}(\mathrm{RELU}(\mathrm{Conv2D}(\mathrm{ReLU}(\mathrm{Conv2D}(\mathbf{X},\mathbf{W_2})),\mathbf{W_3})),\mathbf{W_4})
\end{equation}

where $\mathbf{W_2}$, $\mathbf{W_3}$, and $\mathbf{W_4}$ correspond to $1\times 1$, $3\times 3$
, and $1\times1$ convolutional kernals, respectively.

The skip connection is implemented as:

\begin{equation}
    \mathbf{Y}=F(\mathbf{X})+\mathrm{Conv2D}(\mathbf{X},\mathbf{W_s})
\end{equation}

where $\mathbf{W_s}$ is a $1\times 1$ convolutional kernal used to align the
input dimensions.

After passing through multiple stacked residual blocks, 
the resulting feature map $\mathbf{Z'}$ undergoes global average pooling 
over its spatial dimensions. This operation aggregates global 
information without introducing additional parameters and helps 
prevent overfitting. The formula is:

\begin{equation}
    \mathbf{Z_{pool}} = \frac{1}{H' \times W'} \sum_{i=1}^{H'} \sum_{j=1}^{W'} \mathbf{Z'}_{i,j}
\end{equation}

where $\mathbf{Z_{pool}}$ is the global feature vector with a shape
of $1\times1\times2048$.

\subsubsection{BiLSTM-Based Video Feature Fusion Model} 
Finally, the video features extracted by 
ResNet-50 are preprocessed and fed into a BiLSTM network, which 
hierarchically models both short- and long-range temporal patterns across frames 
across the frame sequence. By 
modeling the sequential structure, the BiLSTM extracts key 
time-dependent features and establishes contextual relationships 
within the video, yielding a temporally aggregated video embedding.

\subsection{Multimodal Audio-Visual Emotion Recognition}
\begin{figure}[h]
    \includegraphics[width=0.98\textwidth]{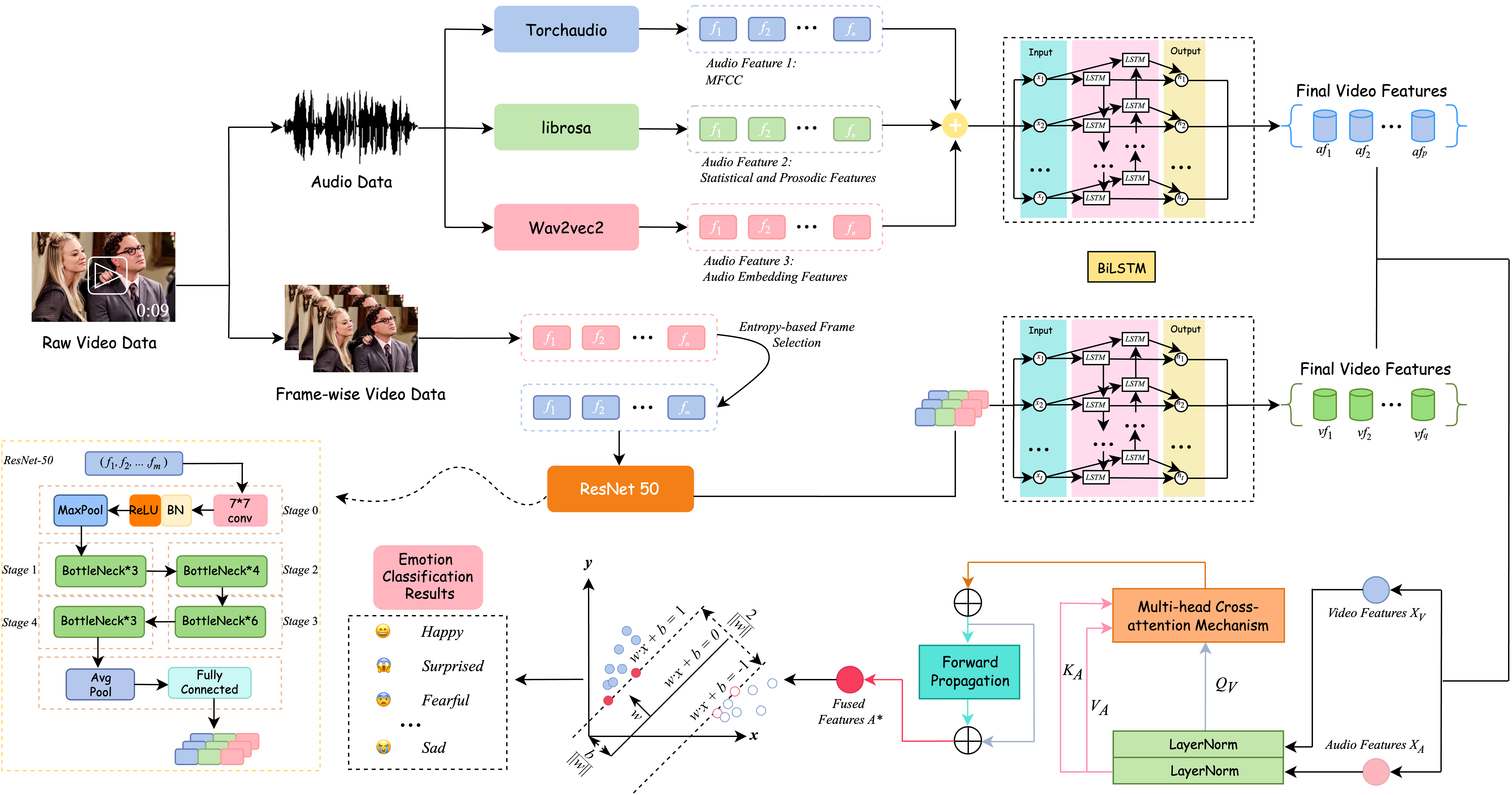}
    \caption{Architecture of the multimodal audio-visual emotion recognition model} \label{fig3}
\end{figure}

We construct a multimodal emotion recognition model by combining 
the two unimodal feature extraction modules proposed in 
Sections~\ref{sec:audio} and~\ref{sec:video} with a feature-level fusion strategy based on a 
multi-head attention mechanism introduced in this section. 
The overall architecture is illustrated in Fig.~\ref{fig3}.

\subsubsection{Feature-Level Fusion Based on Multi-Head Attention Mechanism} 

We adopt a feature-level fusion strategy based on a multi-head 
attention mechanism to perform multimodal emotion recognition.

Specifically, given the audio feature vector $\boldsymbol{X_A}$ and video feature 
vector $\boldsymbol{X_V}$ obtained from the models described in 
Sections~\ref{sec:audio} and~\ref{sec:video}, we first apply a 
LayerNorm layer to $\boldsymbol{X_V}$ to obtain the query vector 
$\boldsymbol{Q_V}$, and apply LayerNorm to $\boldsymbol{X_A}$ to 
obtain the key vector $\boldsymbol{K_A}$ and value vector $\boldsymbol{V_A}$. 
Then, $\boldsymbol{Q_V}$, $\boldsymbol{K_A}$, and $\boldsymbol{V_A}$ are passed 
through a multi-head cross-attention mechanism to compute the output vector 
$\boldsymbol{A_X}$. The attention mechanism is defined as follows:

\begin{equation}
    \mathrm{CrossAttention}(\boldsymbol{X_A},\boldsymbol{X_V})=\mathrm{Softmax}(\frac{\boldsymbol{Q_V}\boldsymbol{K_A^{\top}}}{\sqrt{d_V}})\boldsymbol{V_A}
\end{equation}

Here, $\boldsymbol{X_V}$ denotes the video feature vector, $\boldsymbol{Q_V}$ is 
the video query vector, and $\boldsymbol{X_A}$ is the audio feature vector.
$\boldsymbol{K_A}$ and $\boldsymbol{V_A}$ represent the audio key and value vectors,respectively.
We have $\boldsymbol{X_A} \in \mathbb{R}^{n \times d_A}$, 
$\boldsymbol{X_V} \in \mathbb{R}^{n \times d_V}$, where
$\mathbf{W^Q} \in \mathbb{R}^{d_A \times d_k}$ and 
$\mathbf{W^K} \in \mathbb{R}^{d_V \times d_k}$
are learnable parameter matrices. Here, $\mathbb{R}$ denotes the set of real matrices,
$d_A$ is the dimension of the audio feature vector, $d_V$ is the dimension 
of the value vector, and $d_k$ is the dimension of the key vector.

The output attention vector $\boldsymbol{A_X}$ is added to the query vector 
$\boldsymbol{Q_V}$ and normalized to form an intermediate representation. This is 
passed through a feed-forward layer and added to the previous output to obtain 
the final fused feature vector $\boldsymbol{A^*}$.

This attention-based fusion strategy adaptively calibrates weights to different 
modalities, addressing feature redundancy and complementarity. It effectively 
captures long-range dependencies and contextual relevance while maintaining 
real-time performance. Additionally, it handles data imbalance and improves 
the recognition accuracy for minority emotion classes, enhancing the 
robustness and precision of the overall system.

\subsubsection{Emotion Classification with SVM}

To effectively map the high-dimensional fused features to emotion categories 
and to ensure a fair comparison with several classic baseline methods that 
also utilize SVMs, we employ a Support Vector Machine (SVM) as the final 
classification module. SVMs, known for their strong performance in high-dimensional 
spaces and robustness to overfitting on smaller datasets, provide a powerful 
decision boundary for our extracted features.

In classification tasks, SVM maps the input data into a
high-dimensional feature space and identifies a hyperplane 
that best separates the classes. Given a sample set 
$\boldsymbol{X} = \{\boldsymbol{X}_1, \boldsymbol{X}_2, \dots, \boldsymbol{X}_N\}$ 
and corresponding labels 
$\boldsymbol{y} = \{\boldsymbol{y}_1, \boldsymbol{y}_2, \dots, \boldsymbol{y}_N\}$, 
the decision hyperplane is defined by:

\begin{equation}
    f(x) = \boldsymbol{w}^\top \boldsymbol{X} + b = 0  
\end{equation}

where $\boldsymbol{w}$ is the weight vector perpendicular to the 
hyperplane, and $b$ is the bias. The margin, defined as the distance 
between the closest samples on either side of the 
hyperplane (i.e., the support vectors), is:

\begin{equation}
    d = \frac{2}{\lVert \boldsymbol{w} \rVert}
\end{equation}

Support vectors satisfy $f(x) = 1$ for positive 
and $f(x) = -1$ for negative classes. The 
functional margin of a point $\mathbf{x}$ with label $y$ 
is defined as:

\begin{equation}
    \gamma = y (\boldsymbol{w}^\top \mathbf{x} + b)
\end{equation}

This formulation allows SVM to generalize well even in 
high-dimensional spaces, making it suitable for distinguishing 
emotional categories based on fused feature representations.

\section{Experiments}
\subsection{Evaluation of Single-Modal Emotion Recognition Models}

\subsubsection{Experimental Description}
\label{3_1_1}
To evaluate the effectiveness of our designed single-modal audio 
and video emotion recognition models, we conducted experiments 
on the IEMOCAP and MELD datasets. Both datasets provide strictly synchronized multimodal data, 
enabling independent 
assessment of single-modal recognition models.

{\bfseries MELD (Multimodal EmotionLines Dataset)} is an extension of 
the EmotionLines dataset for multimodal emotion recognition. 
It contains over 13,000 utterances from more than 1,400 dialogue 
segments of the TV show Friends. Each utterance is annotated 
with one of seven emotions: anger, disgust, sadness, joy, neutral, 
surprise, and fear, and includes corresponding audio, visual, 
and textual data. The distribution is shown in Table~\ref{tab1}.

\begin{table}[htbp]
    \centering
    \caption{Distribution Table of the MELD Dataset}
    \label{tab1}
    \begin{tabularx}{\textwidth}{|>{\centering\arraybackslash}X
                               |>{\centering\arraybackslash}X
                               |>{\centering\arraybackslash}X
                               |>{\centering\arraybackslash}X|}
    \hline
    \textbf{Emotion Label} & \textbf{Training Set} & \textbf{Development Set} & \textbf{Test Set} \\
    \hline
    anger & 1109 & 153 & 345 \\
    disgust & 271 & 22 & 68 \\
    fear & 268 & 40 & 50 \\
    joy & 1743 & 163 & 402 \\
    neutral & 4710 & 470 & 1256 \\
    sadness & 683 & 111 & 208 \\
    surprise & 1205 & 150 & 281 \\
    \hline
    \end{tabularx}
\end{table}

{\bfseries IEMOCAP}, released by the USC SAIL lab, includes scripted and 
improvised dyadic interactions by 10 professional actors. During data collection, 
synchronized audio, facial expressions, head movement, and hand gestures were recorded. 
The dataset totals approximately 12 hours and provides fine-grained labels 
(including intensity and duration annotations), enabling multimodal temporal modeling research.

The experiments were conducted on a Linux system using Python 3.11, PyTorch 2.1.2, 
and CUDA 11.8.

We framed the emotion classification task as a seven-class problem (anger, disgust, 
sadness, joy, neutral, surprise, fear). The performance was evaluated using 
Accuracy, Precision, Recall, and F1-Score. Confusion matrices were used for 
visualization in selected experiments.

All models, including the unimodal encoders and the final multimodal network, 
are trained end-to-end to minimize the categorical cross-entropy loss between 
the predicted emotion probabilities and the ground-truth labels.

\subsubsection{Evaluation of Audio Feature Extraction Model}
\begin{figure}[htbp]
    \centering
    \includegraphics[width=0.8\textwidth]{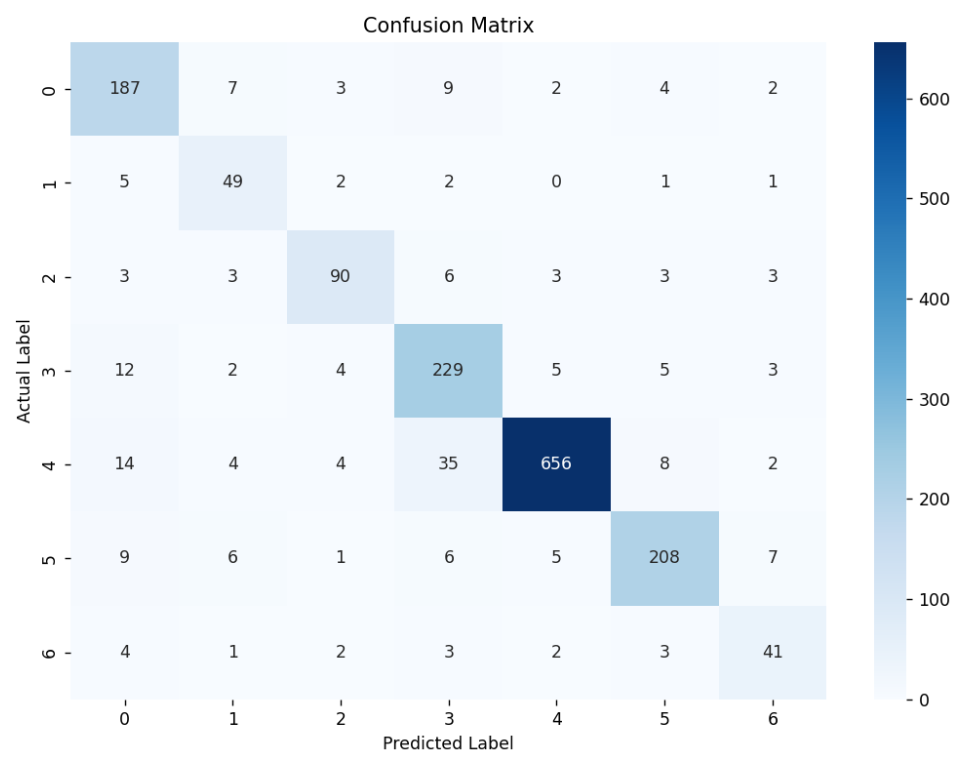}
    \caption{Confusion Matrix of Audio-Based Emotion Recognition} \label{fig4}
\end{figure}

We evaluated the proposed multi-feature audio extraction model 
(as described in section~\ref{sec:audio}) on the MELD dataset. The dataset was split into 
training and test sets in an 8:2 ratio. Fig.~\ref{fig4}. presents the confusion matrix 
results on the test set.

The results show good recognition accuracy for emotions such as anger, joy, and neutral, 
while performance on fear and sadness was relatively weaker. The overall accuracy reached 
80.3\%, indicating strong generalization capability.

To further validate the model$'$s effectiveness, we compared 
it against three commonly used audio feature extraction networks: RNN~\cite{ref_lncs6}, 
2D CNN~\cite{ref_article3}, and 1D CNN-LSTM~\cite{ref_article4}, under the same classifier. 
These established models provide a common baseline for ablative studies, offering 
a clear reference for measuring incremental gains. While newer architectures exist, 
our focus is to demonstrate improvements from our novel multi-feature and 
ResNet-BiLSTM design over these fundamental approaches. All baselines were 
implemented using original reported configurations (see cited works for details).
The accuracy comparison is shown in Table~\ref{tab2}.

\begin{table}[h]
    \centering
    \caption{Comparison of Different Audio Feature Extraction Algorithms}
    \label{tab2}
    \begin{tabularx}{\textwidth}{|>{\centering\arraybackslash}X
                               |>{\centering\arraybackslash}X
                               |>{\centering\arraybackslash}X
                               |}
    \hline
    \textbf{Methods} & \textbf{Training Set} & \textbf{Test Set}  \\
    \hline
    RNN~\cite{ref_lncs6} & 68.4\% & 68.3\%  \\
    2D CNN~\cite{ref_article3} & 69.2\%& 69.7\% \\
    1D CNN-LSTM~\cite{ref_article4} & 73.9\%& 76.5\% \\
    \textbf{Multi-audioFE} & \textbf{78.6\%} & \textbf{80.3\%} \\
    \hline
    \end{tabularx}
\end{table}

Our model achieved notable improvements: training accuracy increased 
by 10.2\%, 9.6\%, and 4.7\% respectively, and test accuracy increased 
by 12\%, 10.6\%, and 3.8\%, demonstrating clear advantages over baseline models 
in audio-based emotion recognition.

\subsubsection{Evaluation of Video Feature Extraction Model} 
For the evaluation of video feature extraction, we first preprocessed the data 
to obtain $256\times256$ standard facial images. These frames were passed into the 
ResNet50-BiLSTM model proposed in Section 2.2 for feature extraction. 
We compared the performance with two classic video models: AlexNet~\cite{ref_lncs7} and 
GoogleNet~\cite{ref_lncs8}. The results are presented in Table~\ref{tab3}.

\begin{table}[h]
    \centering
    \caption{Comparison of Different Video Feature Extraction Algorithms}
    \label{tab3}
    \begin{tabularx}{\textwidth}{|>{\centering\arraybackslash}p{4cm}  
                               |>{\centering\arraybackslash}X
                               |>{\centering\arraybackslash}X
                               |>{\centering\arraybackslash}X
                               |>{\centering\arraybackslash}X|}
    \hline
    \textbf{Methods} & \textbf{Accuracy} & \textbf{Precision} & \textbf{Recall} & \textbf{F1-Score} \\
    \hline
    AlexNet~\cite{ref_lncs7} & 68.1\% & 69.6\% & 66.7\% & 67.9\% \\
    GoogleNet~\cite{ref_lncs8} & 67.3\% & 66.9\% & 66.9\% & 66.7\% \\
    \textbf{ResNet50-BiLSTM} & \textbf{71.7\%} & \textbf{72.3\%} & \textbf{71.4\%} & \textbf{71.8\%} \\
    \hline
    \end{tabularx}
\end{table}

Under the same emotion classifier, AlexNet achieved an average test accuracy 
of 68.1\% on the MELD dataset, while GoogleNet slightly underperformed at 67.3\%. 
Our ResNet50-BiLSTM model, benefiting from residual connections and deep temporal modeling, 
achieved 71.7\% average accuracy.

The results demonstrate that the proposed ResNet50-BiLSTM model is effective in 
capturing spatio-temporal features and yields superior performance in video-based 
emotion recognition.

\subsection{Evaluation of Multimodal Emotion Recognition Model Performance}
To evaluate the effectiveness of our proposed multimodal audio-visual emotion recognition 
model, we conducted experiments on the IEMOCAP and MELD datasets. The details of the datasets 
and experimental environment have already been described in section~\ref{3_1_1} and are omitted 
here. We continued to use Accuracy, Recall, Precision, and F1-Score as evaluation metrics.

\begin{figure}[htbp]
    \centering
    \includegraphics[width=0.8\textwidth]{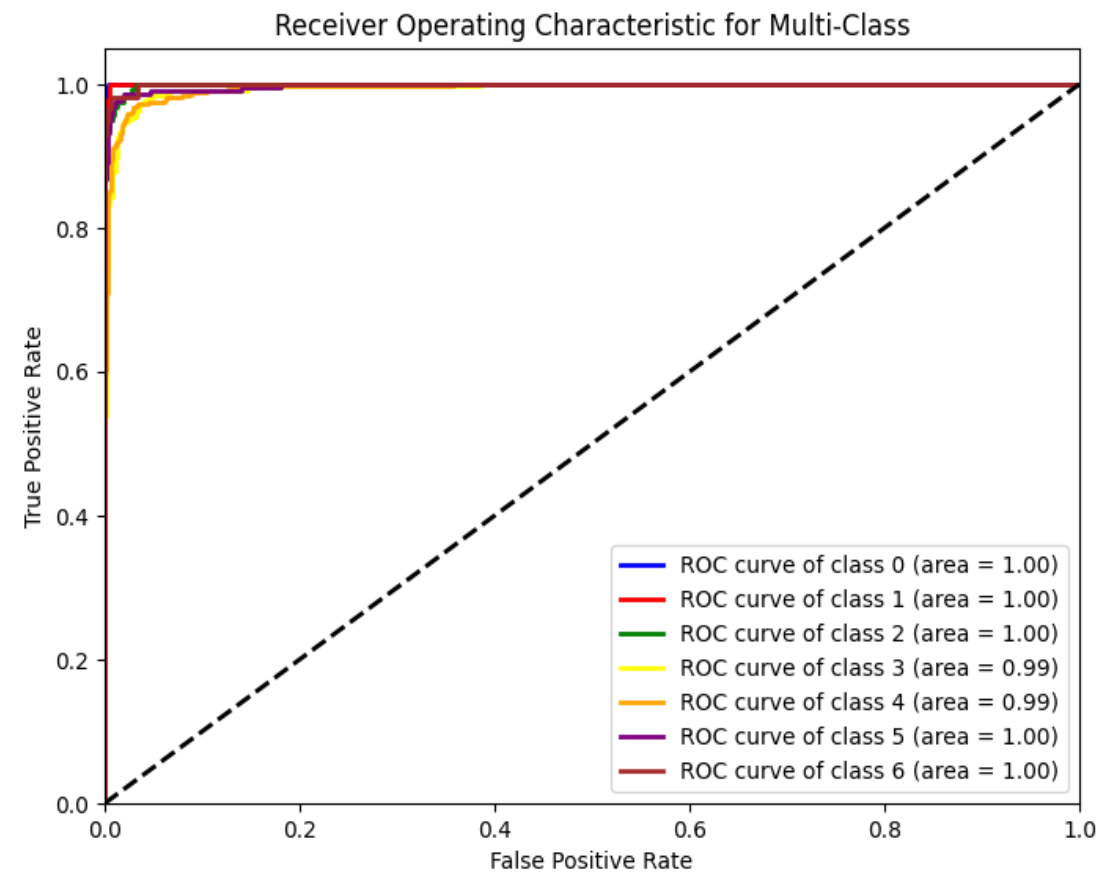}
    \caption{ROC Curves of Multimodal Emotion Recognition} \label{fig5}
\end{figure}

The MELD dataset was divided into training and test sets with an 8:2 ratio. 
The ROC curves for each emotion class on the test set are illustrated in Fig.~\ref{fig5}..

In the figure, each curve represents the ROC performance of an individual emotion category. 
The closer a curve is to the top-left corner, the better the model$'$s recognition ability 
for that emotion. As shown, most curves cluster near the top-left corner, indicating 
excellent classification performance.

The x-axis represents the false positive rate (FPR), which is the proportion of negative 
samples misclassified as positive. All emotion categories exhibit extremely low FPR, 
close to zero, indicating minimal false alarms. The y-axis denotes the true positive 
rate (TPR), with values close to 1, showing that the model correctly identifies nearly 
all positive samples.

Furthermore, the area under the curve (AUC) values for all seven emotion categories 
approach 1.00, further validating the strong performance of our model across all 
classes. These results preliminarily confirm the effectiveness of the proposed multimodal 
emotion recognition model.

\begin{figure}[htbp]
    \centering
    \includegraphics[width=0.8\textwidth]{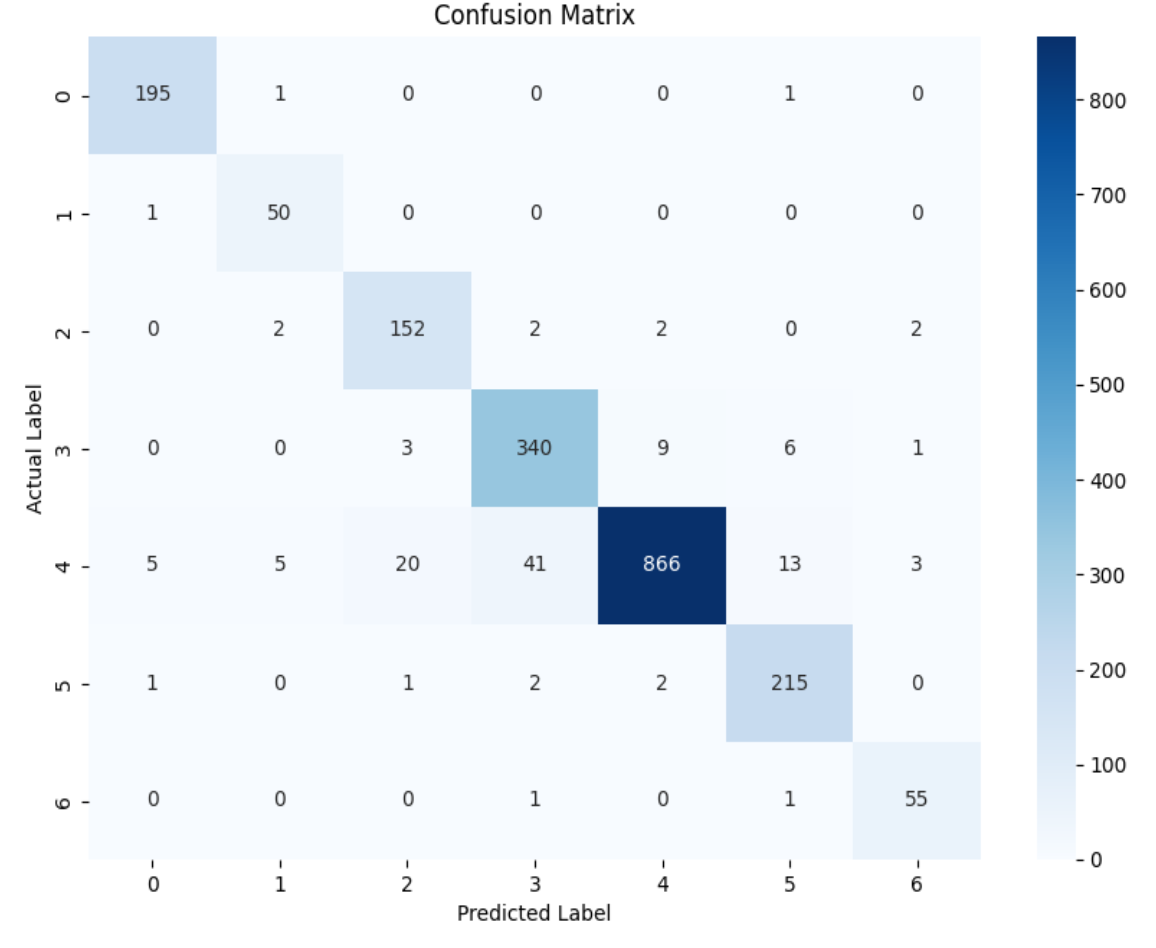}
    \caption{Confusion Matrix of Multimodal Emotion Recognition} \label{fig6}
\end{figure}

Additionally, the confusion matrix of the model's performance on the MELD test set is 
shown in Fig.~\ref{fig6}..

Based on the confusion matrix, we calculated the accuracy, precision, recall, 
and F1-Score for each emotion category, as presented in Table~\ref{tab4}.

\begin{table}[htbp]
    \centering
    \caption{Results Table of Multimodal Emotion Recognition}
    \label{tab4}
    \begin{tabularx}{\textwidth}{|>{\centering\arraybackslash}p{3.8cm}  
                               |>{\centering\arraybackslash}X
                               |>{\centering\arraybackslash}X
                               |>{\centering\arraybackslash}X
                               |>{\centering\arraybackslash}X|}
    \hline
    \textbf{Emotion Label} & \textbf{Accuracy} & \textbf{Precision} & \textbf{Recall} & \textbf{F1-Score} \\
    \hline
    anger & 98.9\% & 96.5\% & 98.9\% & 97.7\% \\
    disgust & 98.0\% & 89.2\% & 98.0\% & 93.4\% \\
    fear & 95.0\% & 85.4\% & 95.0\% & 90.0\% \\
    joy & 94.7\% & 86.0\% & 94.7\% & 90.2\% \\
    neutral & 90.8\% & 97.6\% & 90.8\% & 94.1\% \\
    surprise & 97.2\% & 96.8\% & 97.7\% & 97.2\% \\
    sadness & 96.4\% & 88.7\% & 96.4\% & 92.4\% \\
    micro-average & 93.7\% & 93.7\% & 93.7\% & 93.7\% \\
    \hline
    \end{tabularx}
\end{table}

From the confusion matrix and evaluation metrics, we observe strong classification 
performance across all seven emotion categories, even with limited training samples. 
The average accuracy reaches 93.7\%, and for the large-scale ``neutral'' class, 
accuracy remains above 90\% despite a slight drop.

These results demonstrate the high effectiveness of our model. By introducing an 
attention mechanism that dynamically allocates weights to different modalities, the 
model fully leverages complementary multimodal features, maintaining high classification 
accuracy.

Moreover, on the imbalanced test set, all emotion categories still achieve high accuracy 
and F1 scores, confirming the robustness and generalization ability of our model.

To further assess the generalizability of our approach, we conducted comparative 
experiments on the IEMOCAP dataset against several state-of-the-art multimodal emotion 
recognition models. The models and results are summarized in Table~\ref{tab5}.

\begin{table}[htbp]
    \centering
    \caption{Comparative Results of Multimodal Emotion Recognition Models}
    \label{tab5}
    \begin{tabularx}{\textwidth}{|>{\centering\arraybackslash}X
                               |>{\centering\arraybackslash}p{4cm}
                               |>{\centering\arraybackslash}X
                               |>{\centering\arraybackslash}X|}
    \hline
    \textbf{Model} & \textbf{Fusion Strategy} & \textbf{Accuracy} & \textbf{F1-Score}  \\
    \hline
    EF-LSTM & Feature-level Fusion & 79.8\% & 79.1\% \\
    BERT & Feature-level Fusion & 82.7\% & 83.6\% \\
    CNN-LSTM & Decision-level Fusion & 82.9\% & 84.1\% \\
    \textbf{Ours} & \textbf{Feature-level Fusion} & \textbf{90.3\%} & \textbf{88.5\%} \\
    \hline
    \end{tabularx}
\end{table}

As shown in Table 5, the proposed model, which integrates our audio extractor and 
ResNet50-BiLSTM visual extractor via multi-head attention-based fusion, outperforms 
EF-LSTM, BERT, and CNN-LSTM on the IEMOCAP dataset. Specifically, it achieves 
10.5\%, 7.6\%, and 7.4\% higher accuracy, and 9.4\%, 4.9\%, and 4.5\% higher F1 scores, 
respectively.

These improvements stem from our modeling of intra-modal context, inter-modal 
interactions, and the incorporation of temporal dependencies to simulate the continuity 
of emotional expression.

To validate the effectiveness of our fusion strategy, Table~\ref{tab6} presents the results 
of ablation studies conducted on the MELD dataset.

\begin{table}[htbp]
    \centering
    \caption{Ablation Study Results for Multimodal Emotion Recognition}
    \label{tab6}
    \begin{tabularx}{\textwidth}{|>{\centering\arraybackslash}p{4cm}
                               |>{\centering\arraybackslash}p{4cm}
                               |>{\centering\arraybackslash}X
                               |>{\centering\arraybackslash}X|}
    \hline
    \textbf{Model} & \textbf{Fusion Strategy} & \textbf{Accuracy} & \textbf{F1-Score}  \\
    \hline
    Non-fused Feature Layer & Concatenated Vector & 82.9\% & 84.1\% \\
    Fused Feature Layer & Attention-based Fusion & 93.7\% & 93.5\% \\
    \hline
    \end{tabularx}
\end{table}

The results show that incorporating multi-head attention enables dynamic 
modality weighting and extraction of complementary information. Compared to 
models without fusion, our approach improves accuracy by 10.8\% and F1-Score by 9.4\%.

By integrating unimodal features using attention-based fusion, our model enhances inter-modal 
interactions and classification performance, validating the effectiveness of the proposed 
strategy.

\section{Conclusions}
In this paper, we proposed a multimodal emotion recognition framework that integrates 
deep feature extraction from both audio and video modalities with an attention-based 
fusion mechanism. The aim was to address the limitations of existing unimodal models 
in extracting high-quality features and to enhance the synergy between different 
modalities for improved classification performance.

Specifically, we designed a multi-feature audio extraction model that incor- porates 
semantic embeddings from Wav2Vec2, MFCC features, and statistical acoustic descriptors. 
This design enables the model to capture audio cues from multiple perspectives, improving 
the robustness and expressiveness of emotional information. For visual features, 
we introduced a ResNet50-BiLSTM architec- ture that combines residual learning and 
temporal modeling to effectively extract rich emotional representations from facial video 
frames.

To bridge the gap between modalities, we implemented a multi-head attention-based 
feature-level fusion strategy. This fusion approach dynamically allocates attention 
weights across modalities, preserves the complementarity of multimodal data, and 
mitigates the risk of information loss during fusion. Ex- perimental results on the 
MELD and IEMOCAP datasets demonstrate that our proposed framework achieves competitive 
performance across all major metrics and outperforms several state-of-the-art baselines.

Although the proposed method achieves promising experimental results, there remains room 
for further improvement. First, the current work primarily focuses on audio and visual 
modalities; future research could incorporate physiological signals such as EEG, skin 
conductance, and heart rate to better capture emotional states. For instance, Pardhan et al. 
combined DenseNet and LSTM to perform multimodal emotion recognition using physiological 
data~\cite{ref_article5}. Second, with the advancement of large multimodal models, pretrained networks 
offer a potential avenue to enhance feature extraction. Minoo Shayaninasab et al. have 
explored multimodal emotion recognition using Transformer-based pretrained models~\cite{ref_lncs9}. 
Finally, the Transformer architecture shows great promise in cross-modal modeling, 
enabling higher-level fusion strategies that improve generalizability and interpretability. 
For example, Zaidi et al. proposed a multimodal dual-attention Transformer for emotion 
recognition, capturing rich inter-modal emotional interactions~\cite{ref_lncs10}.


\begin{thebibliography}{15}

    \bibitem{ref_article1}
    Pang, J.-H., Hou, Z.-P., Li, Z.-N., et al.: A Survey of Multimodal Emotion Recognition Research. J. Intell. Syst. \textbf{15}(4), 633--645 (2020). \doi{10.11992/tis.202001032}
    
    \bibitem{ref_article2}
    Salas-Cáceres, J., Lorenzo-Navarro, J., Freire-Obregón, D. et al. Multimodal emotion recognition based on a fusion of audiovisual information with temporal dynamics. Multimed Tools Appl (2024). https://doi.org/10.1007/s11042-024-20227-6

    \bibitem{ref_lncs1}
    Kova\v{c}evi\'{c}, N., Holz, C., Gross, M., Wampfler, R.: On Multimodal Emotion Recognition for Human-Chatbot Interaction in the Wild. In: Proc. Int. Conf. on Multimodal Interaction (ICMI 2024), pp. 1--12. ACM, San Jose (2024). \doi{10.1145/3678957.3685759}

    \bibitem{ref_lncs2}
    Li, Y., Sun, Q., Murthy, S.M.K., Alturki, E., Schuller, B.W.: GatedxLSTM: A Multimodal Affective Computing Approach for Emotion Recognition in Conversations. arXiv preprint arXiv:2503.20919 (2025). \url{https://arxiv.org/abs/2503.20919}

    \bibitem{ref_lncs3}
    Zhao, S., Ma, Y., Gu, Y., Yang, J., Xing, T., Xu, P., Hu, R., Chai, H., Keutzer, K.: An End-to-End Visual-Audio Attention Network for Emotion Recognition in User-Generated Videos. In: Proc. 34th AAAI Conf. on Artificial Intelligence (AAAI 2020), pp. 303--311. AAAI Press, New York (2020)
    
    \bibitem{ref_lncs4}
    Avro, S.B.H., Taher, T., Mamun, N.: EmoTech: A Multi-modal Speech Emotion Recognition Using Multi-source Low-level Information with Hybrid Recurrent Network. arXiv preprint arXiv:2501.12674 (2025). \url{https://arxiv.org/abs/2501.12674}

    \bibitem{ref_lncs5}
    Dai, W., Zheng, D., Yu, F., Zhang, Y., Hou, Y.: A Novel Approach to Multimodal Emotion Recognition: Multimodal Semantic Information Fusion. arXiv preprint arXiv:2502.08573 (2025). \url{https://arxiv.org/abs/2502.08573}

    \bibitem{ref_lncs6}
    Tzinis, E., Potamianos, A.: Segment-Based Speech Emotion Recognition Using Recurrent Neural Networks. In: Proc. 7th Int. Conf. on Affective Computing and Intelligent Interaction (ACII), pp. 190--195. IEEE, San Antonio (2017)
    
    \bibitem{ref_article3}
    Mocanu, B., Tapu, R., Zaharia, T.: Multimodal Emotion Recognition Using Cross-Modal Audio-Video Fusion with Attention and Deep Metric Learning. Image Vis. Comput. \textbf{132}, 104664 (2023)
    
    \bibitem{ref_article4}
    Moon, E., Sagar, A.S.M.S., Kim, H.S.: Multimodal Daily-Life Emotional Recognition Using Heart Rate and Speech Data from Wearables. IEEE Access \textbf{12}, 1--10 (2024). \url{https://ieeexplore.ieee.org/document/10402679} (Accessed: Jan. 27, 2025)
    
    \bibitem{ref_lncs7}
    Krizhevsky, A., Sutskever, I., Hinton, G.: ImageNet Classification with Deep Convolutional Neural Networks. In: Adv. Neural Inf. Process. Syst. (NeurIPS), vol. 25, pp. 1097--1105. Curran Associates Inc., Red Hook (2012)
    
    \bibitem{ref_lncs8}
    Szegedy, C., Liu, W., Jia, Y., et al.: Going Deeper with Convolutions. In: Proc. IEEE Conf. Comput. Vis. Pattern Recognit. (CVPR), pp. 1--9. IEEE, Boston (2015)
    
    \bibitem{ref_article5}
    Pradhan, A., Srivastava, S.: Hybrid Densenet with Long Short-Term Memory Model for Multi-modal Emotion Recognition from Physiological Signals. Multimed. Tools Appl. \textbf{83}, 35221--35251 (2024). \doi{10.1007/s11042-023-16933-2}

    \bibitem{ref_lncs9}
    Shayaninasab, M., Babaali, B.: Multi-Modal Emotion Recognition by Text, Speech and Video Using Pretrained Transformers. arXiv preprint arXiv:2402.07327 (2024). \url{https://arxiv.org/abs/2402.07327}

    \bibitem{ref_lncs10}
    Zaidi, S.A.M., Latif, S., Qadir, J.: Cross-Language Speech Emotion Recognition Using Multimodal Dual Attention Transformers. arXiv preprint arXiv:2306.13804 (2023). \url{https://arxiv.org/abs/2306.13804}

\end{thebibliography}
\end{document}